\documentclass[10pt,twocolumn,letterpaper]{article}

\usepackage[pagenumbers]{iccv} 

\usepackage{multirow} 
\usepackage{colortbl} 

\usepackage{pgfplots} 
\usepackage{amssymb} 

\definecolor{iccvblue}{rgb}{0.21,0.49,0.74}
\usepackage[pagebackref,breaklinks,colorlinks,allcolors=iccvblue]{hyperref}

\def\paperID{5777} 
\def\confName{ICCV}
\def\confYear{2025}

\title{RT-DATR: Real-time Unsupervised Domain Adaptive Detection Transformer with Adversarial Feature Alignment}

\author{Feng Lv\\
Baidu Inc\\
Beijing\\
{\tt\small lvfeng02@baidu.com}
\and
Guoqing Li\\
Southeast University \\
Nanjing \\
{\tt\small li\_guoqing@aa.seu.edu.cn}
\and
Jin Li\\
Shaanxi Normal University \\
Shaanxi \\
{\tt\small j.lixjtu@gmail.com}
\and
Chunlong Xia\thanks{corresponding author} \\
Baidu Inc \\
Beijing \\
{\tt\small xiachunlong@baidu.com}
}

\begin{document}
\maketitle


\begin{abstract}
Despite domain-adaptive object detectors based on CNN and transformers have made significant progress in cross-domain detection tasks, it is regrettable that domain adaptation for real-time transformer-based detectors has not yet been explored. Directly applying existing domain adaptation algorithms has proven to be suboptimal. In this paper, we propose RT-DATR, a simple and efficient real-time domain adaptive detection transformer. Building on RT-DETR as our base detector, we first introduce a local object-level feature alignment module to significantly enhance the feature representation of domain invariance during object transfer. Additionally, we introduce a scene semantic feature alignment module designed to boost cross-domain detection performance by aligning scene semantic features. Finally, we introduced a domain query and decoupled it from the object query to further align the instance feature distribution within the decoder layer, reduce the domain gap, and maintain discriminative ability. Experimental results on various cross-domian benchmarks demonstrate that our method outperforms current state-of-the-art approaches. Code is available at \href{https://github.com/Jeremy-lf/RT-DATR}{https://github.com/Jeremy-lf/RT-DATR}.
\end{abstract}

\begin{figure}[ht]
\begin{center}
\centerline{\includegraphics[scale=0.42]{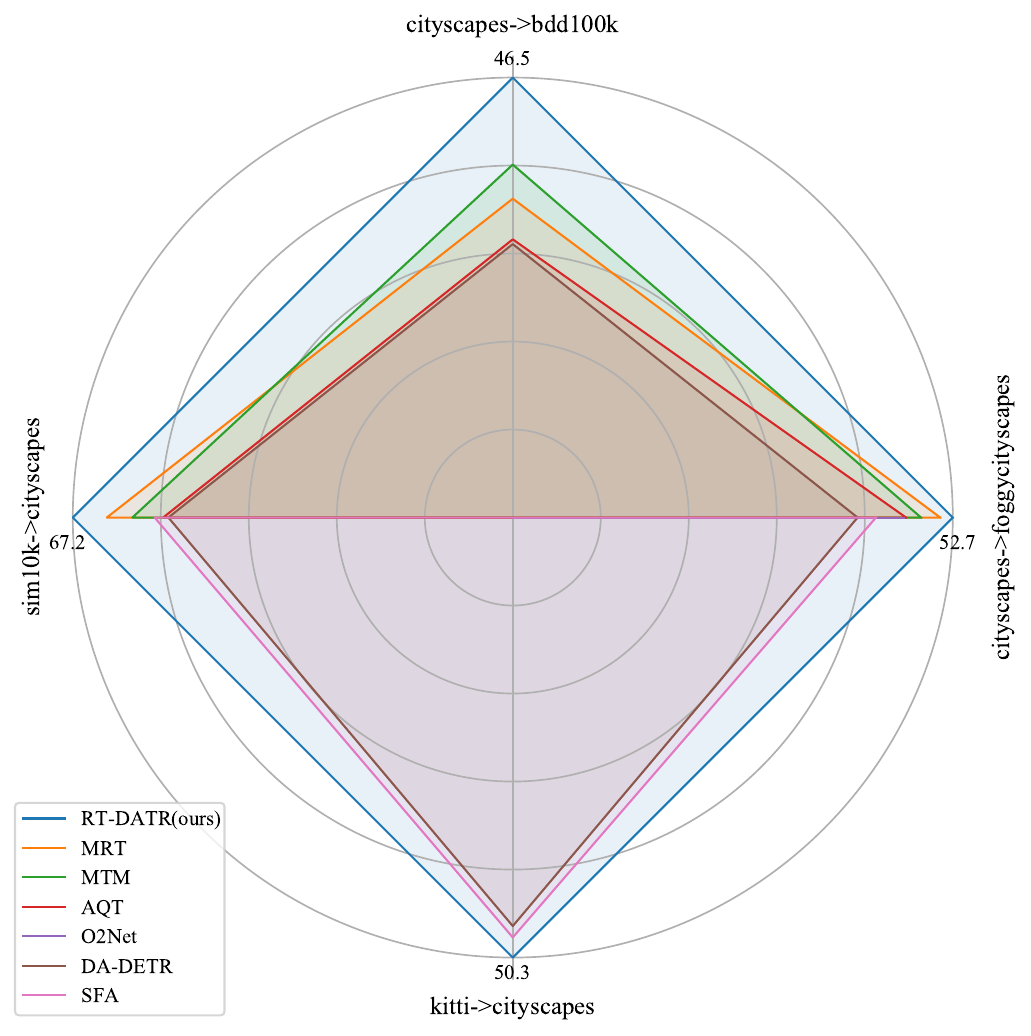}}
\caption{Performance comparison of different benchmarks.}
\label{benchmark}
\end{center}
\end{figure}

\section{Introduction}
Object detection is a fundamental task in computer vision, focusing on the localization and classification of objects within images. It has diverse applications across various domains, including autonomous driving, security surveillance, and industrial production. However, most detectors heavily depend on labeled training data, making them vulnerable to performance drops when the testing environment differs from the training conditions. Factors such as variations in weather, camera angles, and cross-city scenarios can significantly impact their effectiveness. This phenomenon, where performance degradation arises due to discrepancies between the distributions of training and testing data, is commonly referred to as domain shift.

To tackle this challenge, an increasing body of research has focused on developing unsupervised domain adaptation (UDA)~\cite{long2016unsupervised} algorithms for object detection. These efforts include adversarial feature learning, image translation, and semi-supervised approaches based on the student-teacher framework. The primary goal of these methods is to leverage both labeled source domain data and unlabeled target domain data to train a detector that can generalize effectively in the target domain. For example, several influential studies integrate adversarial learning with object detection by introducing a domain classifier and a gradient reversal layer (GRL)~\cite{ganin2015unsupervised} into the detector. This approach minimizes the classifier loss while maximizing the detector loss, thereby encouraging the learning of domain-invariant features and enhancing detection performance in the target domain~\cite{chen2018da-faster}. Additionally, methods based on the mean-teacher framework~\cite{deng2021unbiased} explore learning strategies through the interaction between teacher and student models, progressively adapting the object detector to cross-domain detection tasks.

Many existing domain adaptation methods are meticulously designed for CNN-based detectors~\cite{tian2020fcos, ren2016faster-rcnn}. However, with the outstanding performance of transformers in object detection~\cite{carion2020detr}, transformer-based domain-adaptive object detection has gained significant attention and achieved remarkable progress. Unlike CNN-based detectors, transformer-based detectors address domain shifts by aligning feature distributions across various modules. This is accomplished through the integration of domain-specific queries or by leveraging text-to-image models to synthesize data resembling the target domain from the source domain~\cite{huang2024blenda}. These approaches progressively reduce the domain gap, enabling superior performance in the target domain. 

Nevertheless, domain-adaptive detection transformers face several challenges. On the one hand, their high computational complexity and suboptimal performance in cross-domain detection significantly limit the applicability of these models in real-world scenarios. Although real-time detectors based on transformers have been widely applied in various fields, their cross-domain effectiveness remains underexplored. On the other hand, existing domain adaptation methods often heavily rely on specific detector architectures. This not only makes it difficult to directly transfer these methods to other detectors but may also increase the model’s inference latency, thereby failing to meet the demands of practical applications.

To address these challenges, in this paper, we propose RT-DATR, which, to our knowledge, is the first real-time domain adaptive detection transformer. As shown in Figure~\ref{benchmark}, it achieved state-of-the-art performance on different benchmarks. We chose RT-DETR~\cite{zhao2024rt-detr} as our base detector because it is currently the most popular and widely used DETR-style real-time detector. The modules we introduced do not add any extra inference latency, ensuring the high performance of the model. Specifically, we propose a local object-level feature alignment (LOFA) module and a scene semantic feature alignment (SSFA) module. The former can achieve pixel-level local alignment based on multi-scale features while also enhancing the feature alignment of the object regions. The latter globally aligns the encoded high-level scene semantic features, enabling the learning of transferable domain-invariant features. We introduced a domain query and decoupled it from the object query, aiming to alleviate the domain gap through adversarial learning while maintaining discriminative ability during prediction. Additionally, we enhanced the model's robustness by incorporating consistency loss in the Instance Feature Alignment (IFA) module. Our main contributions can be summarized as follows:

\begin{itemize}
    \item We proposed RT-DATR, to the best of our knowledge, the first real-time domain-adaptive detection transformer based on RT-DETR. Its real-time capability combined with strong cross-domain detection performance makes it highly valuable for practical applications.
    \item We developed three distinct adversarial feature learning modules, which include components for aligning local object-level features and scene semantic features. These modules leverage adversarial learning both at the object level and globally to acquire domain-invariant features and enhance object transferability. Furthermore, we incorporate domain queries and consistency loss to further minimize domain gaps and enhance the model’s robustness. Notably, these modules are all inference-free.
    \item We conducted extensive experiments and compared them with current state-of-the-art methods. RT-DATR can achieve better object detection performance on adaptive benchmarks in multiple widely studied fields.
\end{itemize}

\section{Related Work}
\subsection{Transformer-based Object Detection}
DETR~\cite{carion2020detr} is the first end-to-end object detector based on transformer. It achieves one-to-one set prediction by using the Hungarian matching algorithm without relying on components such as anchor boxes and non-maximum suppression. Although DETR~\cite{carion2020detr} has the advantage of end-to-end prediction, it suffers from slow convergence and high computational costs. To address these issues, Deformable DETR~\cite{zhu2020deformable-detr} introduces a deformable attention module that focuses attention around the reference point, significantly reducing computation cost and accelerating training convergence.

DINO~\cite{zhang2022dino} employs a mixed query selection strategy for better query initialization and introduces a “look forward twice” scheme to refine box predictions. RT-DETR~\cite{zhao2024rt-detr} is the first  transformer-based real-time detector that achieves significant speed improvements while maintaining high accuracy through efficient hybrid encoder design and minimal uncertainty query selection. RT-DETRv2~\cite{lv2024rt-detrv2} introduces a series of ``bag-of-freebies" to improve flexibility and practicality while optimizing training strategies to boost performance. RT-DETRv3~\cite{wang2024rt-detrv3} proposes a hierarchical dense auxiliary supervision scheme to further enhance detection accuracy and convergence speed. In this paper, we selected RT-DETR as our base detector because it is currently the most popular DETR-style real-time detector. To ensure the efficiency of RT-DATR, we designed the modules to enhance cross-domain detection performance without introducing additional inference latency.

\subsection{UDA for Object Detection}
The goal of unsupervised domain-adaptive object detection is to bridge the domain gap between the source and target domains, thereby improving the detector's performance on target domain data. In this work, we conduct a comprehensive review of existing research and categorize it based on domain adaptation algorithms built on different foundational detectors, including CNN-based and transformer-based.

\noindent \textbf{UDA for CNN-based Detector.} Unsupervised domain-adaptive object detection was first proposed in ~\cite{chen2018da-faster}, based on Faster RCNN~\cite{ren2016faster-rcnn} using two gradient reverse layer~\cite{ganin2015unsupervised} for optimization through minimum-maximum adversarial learning, achieving image-level and instance-level alignment. \cite{saito2019strong} proposed a strong-weak distribution alignment method, which focuses the adversarial alignment loss on images that are globally similar and puts less emphasis on aligning images that are globally dissimilar, achieving significant results. Considering that using domain classifiers to reduce the differences between the source and target domains can ignore the difficulties of these transferable features in processing classification and localization subtasks in object detection.Therefore, task-specific inconsistency alignment~\cite{zhao2022tia} is proposed, by developing a new alignment mechanism in separate task spaces,improving the performance of the detector on both sub-tasks. 

Unlike adversarial feature learning, there are also some works that maximize the data information of the target domain by fully utilizing pseudo-label self-training. For example, PT~\cite{chen2022learning} aims to capture the uncertainty of unlabeled target data from evolving teachers and guide students' learning in a mutually beneficial manner, thereby avoiding the negative impact of pseudo box quality. CMT~\cite{cao2023CMT} proposed a framework that combines contrastive learning and mean teacher self-training, using pseudo labels to extract object level features and optimizing them through contrastive learning. Additionally, methods involving image transformation have been employed to reduce the domain gap. For instance, ConfMix\cite{mattolin2023confmix} is the first approach to utilize a sample mixing strategy based on region-level detection confidence for adaptive object detector learning, adopting YOLOv5\cite{jocher2022yolov5} as its base detector.

\noindent \textbf{UDA for Transformer-based Detector.} However, unlike CNN-based domain adaptive object detection, research on adopting DETR-style architectures for domain adaptive detection tasks remains relatively scarce. SFA\cite{wang2021sfa} proposes a method for sequence feature alignment by introducing domain queries to align the features of the encoder and decoder, respectively, achieving promising results. O$^2$Net\cite{gong2022o2net} improves performance by mitigating the domain shift on the backbone and aligning the output features of the encoder. AQT~\cite{huang2022aqt} further integrates adversarial feature alignment methods into the detection transformer and implements space, channel, and instance-level feature alignment to produce domain-invariant features. BiADT\cite{he2023bidirectional} proposes a novel deformable attention and self-attention approach that aims to achieve bidirectional domain alignment.
DA-DETR\cite{zhang2023da-detr} achieves effective transformation from labeled source domain to unlabeled target domain by fusing CNN and Transformer information. MRT~\cite{zhao2023mrt} masks the multi-scale feature map of the target image through the encoder and auxiliary decoder of the student model, and reconstructs the features to help the student model capture target domain features. 
MTM~\cite{weng2024mean} proposes masked feature alignment methods to alleviate domain shift in a more robust way.

Some approaches adopt image-to-image translation methods, utilizing text-to-image generation models to produce more diverse data from the target domain, thereby enhancing the model's generalization ability. Among them, Blenda~\cite{huang2024blenda} performs adaptive training by generating pseudo-samples from an intermediate domain along with their corresponding soft domain labels. Inspired by the above work, we explore for the first time the combination of adversarial feature learning strategies with a real-time transformer-based detector to enhance cross-domain detection performance.

\begin{figure*}[t]
\centering
  \includegraphics[scale=0.52]{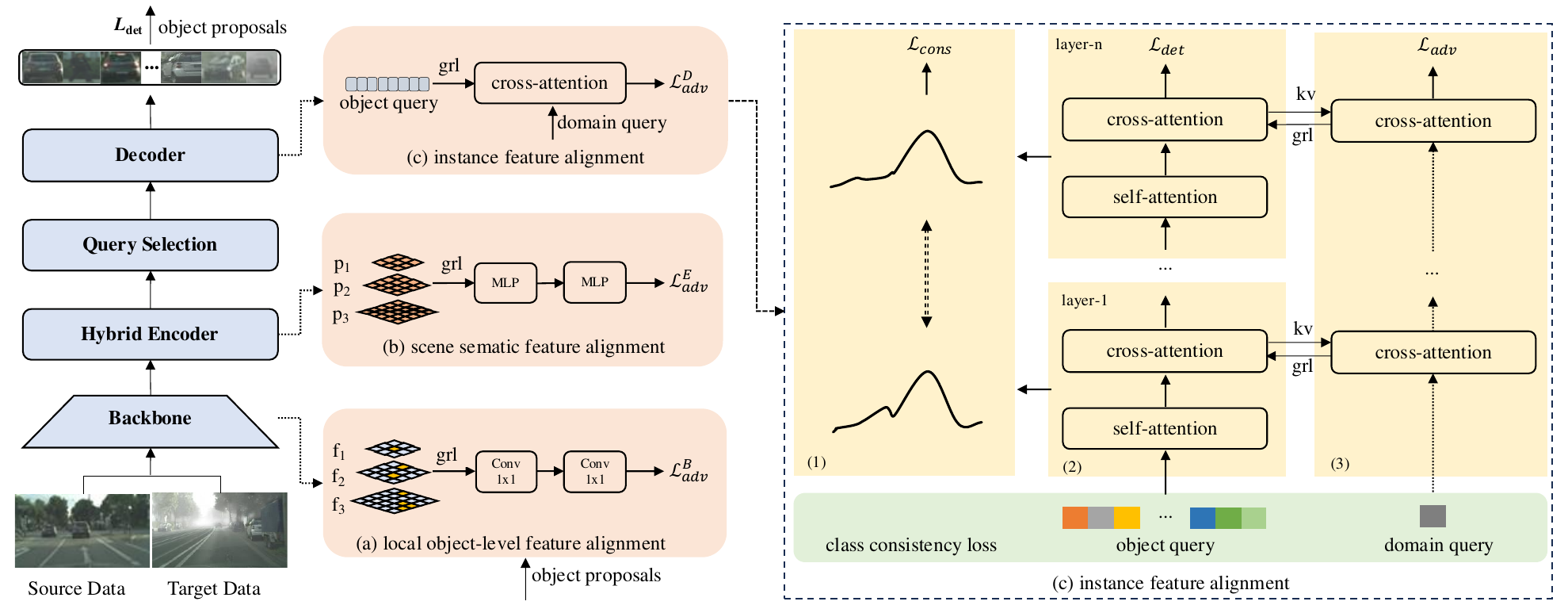}
\caption{The architecture of RT-DATR. It consists of the base detector RT-DETR, along with three feature alignment modules: local object-level, scene semantic, and instance feature alignment modules. Object proposals are the object regions predicted by the model for object-level feature alignment.}
\label{framework}
\end{figure*}

\section{Method}
\subsection{Task Defination and Overview }
It is noteworthy that our focus is not on developing a new real-time object detector. Instead, our efforts are dedicated to improving the cross-domain detection performance of an existing transformer-based real-time detector, and this represents the first research endeavor in this field.

In unsupervised domain adaptation, the training data includes labeled source  images $D_s$, denoted as $\{x_s, y_s\}$, and unlabeled target images $D_t$, denoted as $\{x_t\}$. Our goal is to train an object detector that performs well on target domain data. The overall architecture of our proposed method, RT-DATR, is illustrated in Figure~\ref{framework}. RT-DETR serves as our base detector, with three additional distinct feature alignment modules introduced for domain adaptation. In supervised learning, source domain images are fed into the model, and constraints are imposed by computing the loss between the predictions of RT-DETR and the ground truth, denoted as $\mathcal{L}_{det}$. For unsupervised learning, both source and target domain data are used as inputs, and optimization is performed through the LOFA, SSFA, and IFA modules to drive adversarial fetaure alignment, denoted as $\mathcal{L}_{adv}$.

\subsection{Local and Object-level Feature Alignment}

Adversarial feature alignment within different detector components has proven to be a simple yet effective approach for domain adaptation. However, studies reveal that directly aligning CNN backbone features in transformer-based detectors often degrades performance~\cite{wang2021sfa}. Subsequent research has thus shifted focus to aligning feature distributions in other modules or extracting domain-specific features through domain queries~\cite{huang2022aqt}. In contrast, we emphasize the critical role of the backbone network as the primary module for learning domain-invariant features. The multi-scale features extracted by the backbone capture rich local information, such as textures and colors, which are essential for domain adaptation. To preserve this information, we adopt local feature alignment instead of global image-level alignment, ensuring robust feature matching. Furthermore, we enhance foreground object-level alignment, improving object transferability and boosting detection performance in cross-domain tasks.

In Figure~\ref{framework}a, we illustrate how data from both the source and target domains are processed through a backbone network to extract multi-scale features denoted as ${f_{1}, f_{2}, f_{3}}$. Following this, a gradient reversal layer is integrated at each feature level, and a domain discriminator, which includes a $Conv$1x1 layer, is employed to derive the transformed multi-scale features ${f}'_{1}, {f}'_{2},$ and ${f}'_{3}$. Subsequently, domain classification is performed on each pixel, with the source domain labeled as 0 and the target domain as 1. The optimization objective for aligning local features is defined as follows:

\vspace{-2mm}
\begin{equation}
\begin{split}
    \mathcal{L}_{loc}=\sum_{i=1}^{M}-dlog(D_{b}(f^{s(u,v)}_{i}))+
    \\
    (1-d)log(1-D_{b}(f^{s(u,v)}_{i}))
\end{split}
\end{equation}
Where \(d\) denotes the domain label, \(D_b\) represents the domain discriminator of the backbone, \(f_i\) stands for the \(i\)-th layer among the multi-scale features (\(i = 1, 2, ..., M\)), and \(s(u, v)\) indicates the positional information within the feature layer.

Furthermore, we aim to strengthen the object-level feature alignment within the backbone, which can substantially enhance the transferability of objects in cross-domain scenarios. We assign different classification loss weights based on the position of each pixel in the feature map. Specifically, the classification loss weight is set to 1.5 for pixels within object regions and 0 for those outside.

For source domain images, effective object proposals are directly derived by mapping ground truth bounding boxes to the backbone's feature layers. For target domain images, effective object regions are identified by projecting the predicted bounding boxes from the decoder onto the backbone's multi-scale feature layers, retaining only those with a confidence score exceeding 0.5. Our approach prioritizes feature alignment within object regions while disregarding specific category information, thereby mitigating the error propagation commonly associated with pseudo-labels. The complete optimization process for object-level alignment is formulated as follows:

\begin{equation}
\begin{split}
    \mathcal{L}_{obj}=\sum_{i=1}^{M}-w_{i}dlog(D_{b}(f^{s(u,v)}_{i}))+ 
    \\ w_{i}(1-d)log(1-D_{b}(f^{s(u,v)}_{i})) 
\end{split}
\end{equation}

where $w_{i}$ represent a mask matrix of the same size as the current feature map $f_{i}$, the regions containing objects are assigned a value of 1.5, and the remaining regions are assigned a value of 0. Finally, the overall loss of the backbone network can be described as:
\begin{equation}
    \mathcal{L}^{B}_{adv}= \mathcal{L}_{loc}+\mathcal{L}_{obj} 
\end{equation}
\subsection{Scene Semantic Feature Alignment}

The detection transformer architecture is generally composed of three key modules: the backbone, encoder, and decoder. While CNN-based backbones primarily extract localized and low-level semantic features, transformers leverage the self-attention mechanism to capture long-range dependencies. As the network depth increases, the encoder progressively encodes richer scene-level semantic information. To address domain shift, we employ a scene semantic feature alignment strategy that facilitates the learning of domain-invariant features, thereby enhancing cross-domain generalization.

Specifically, as depicted in Figure~\ref{framework}b, the encoder generates multi-scale features $p_{1}, p_{2},$ and $p_{3}$, which are then used by a discriminator composed of $MLP$ layers to distinguish between domains. 
Ultimately, the model learns to represent domain-invariant features by minimizing the discriminator’s loss while maximizing the feature error at the encoder layer using GRL. The optimization process can be expressed as follows:
\begin{equation}
    \mathcal{L}^{E}_{adv}= \frac{1}{L} \sum_{j=0}^{L}(-dlog(D_{e}(p_{j})) + (1-d)log(1-D_{e}(p_{j})))
\end{equation}
Where \(p_{j}\) represents the \(j\)-th layer feature (with \(j=0,1,...,L\)) output by the encoder, and \(D_{e}\) denotes the discriminator within the encoder.

\begin{table*}[ht!]
\centering
\setlength{\tabcolsep}{5.0 pt} 
\renewcommand{\arraystretch}{1.2}
\begin{tabular}{l|c|ccccccccc}
\hline
Method                                & Backbone & person        & rider         & car           & truck         & bus           & train         & mcycle        & bicycle       & mAP           \\ \hline
Source Only (Faster R-CNN)            & R50      & 22.4          & 26.6          & 28.5          & 9.0           & 16.0          & 4.3           & 15.2          & 25.3          & 18.4          \\
TIA~\cite{zhao2022tia}                & R50      & 34.8          & 46.3          & 49.7          & 31.1          & 52.1          & 48.6          & 37.7          & 38.1          & 42.3          \\
UMT~\cite{deng2021umt}                & R50      & 33.0          & 46.7          & 48.6          & 34.1          & 56.5          & 46.8          & 30.4          & 37.3          & 41.7          \\
CMT~\cite{cao2023CMT}                 & R50      & 45.9          & 55.7          & 63.7          & 39.6          & 66.0          & 38.8          & 41.4          & 51.2          & 50.3          \\
CAT\cite{kennerley2024cat}            & R50      & 44.6          & 57.1          & 63.7          & 40.8          & \textbf{66.0} & 49.7          & 44.9          & \textbf{53.0} & 52.5          \\ \hline
Source Only (FCOS)                    & R50      & 36.9          & 36.3          & 44.1          & 18.6          & 29.3          & 8.4           & 20.3          & 31.9          & 28.2          \\
OADA~\cite{yoo2022oada}               & R50      & 47.8          & 46.5          & 62.9          & 32.1          & 48.5          & 50.9          & 34.3          & 39.8          & 45.4          \\
HT~\cite{deng2023ht}                  & R50      & 52.1          & 55.8          & 67.5          & 32.7          & 55.9          & 49.1          & 40.1          & 50.3          & 50.4          \\ \hline
Source Only (YOLOv5)                 & CSP-D53      & 34.8          & 37.6          & 48.7          & 14.3          & 30.1          & 8.8          & 14.6          & 28.1          & 27.1          \\ 
ConfMix~\cite{mattolin2023confmix}                  & CSP-D53      & 45.0          & 43.4          & 63.6          & 27.3          & 45.8          & 40.0          & 28.6          & 33.5          & 40.8          \\ \hline
Source Only (Deformable DETR)         & R50      & 37.7          & 39.1          & 44.2          & 17.2          & 26.8          & 5.8           & 21.6          & 35.5          & 28.5          \\
SFA~\cite{wang2021sfa}                & R50      & 46.5          & 48.6          & 62.6          & 25.1          & 46.2          & 29.4          & 28.3          & 44.0          & 41.3          \\
MTTrans~\cite{yu2022mttrans}          & R50      & 47.7          & 49.9          & 65.2          & 25.8          & 45.9          & 33.8          & 32.6          & 46.5          & 43.4          \\
DA-DETR~\cite{zhang2023da-detr}       & R50      & 49.9          & 50.0          & 63.1          & 24.0          & 45.8          & 37.5          & 31.6          & 46.3          & 43.5          \\
O2Net~\cite{gong2022o2net}            & R50      & 48.7          & 51.5          & 63.6          & 31.1          & 47.6          & 47.8          & 38.0          & 45.9          & 46.8          \\
AQT~\cite{huang2022aqt}               & R50      & 49.3          & 52.3          & 64.4          & 27.7          & 53.7          & 46.5          & 36.0          & 46.4          & 47.1          \\
MRT~\cite{zhao2023mrt}                & R50      & \textbf{52.8} & 51.7          & 68.7          & 35.9          & 58.1          & \textbf{54.5} & 41.0          & 47.1          & 51.2          \\
MTM~\cite{weng2024mean}                           & R50      & 51.0          & 53.4          & 67.2          & 37.2          & 54.4          & 41.6          & 38.4          & 47.7          & 48.9          \\ \hline
Source Only (RT-DETR)                 & R34      & 39.2          & 47.7          & 51.6          & 28.8          & 41.2          & 7.2           & 27.3          & 41.7          & 35.6          \\
Source Only (RT-DETR)                 & R50      & 41.4          & 49.7          & 58.0          & 31.5          & 53.0          & 25.5          & 36.0          & 45.3          & 42.6          \\
\rowcolor[HTML]{ECF4FF} RT-DATR(Ours) & R34      & 47.6          & 55.9          & 66.1          & 37.1          & 58.7          & 43.4          & 42.0          & 48.0          & 50.4          \\
\rowcolor[HTML]{ECF4FF} RT-DATR(Ours) & R50      & 49.9          & \textbf{59.7} & \textbf{69.2} & \textbf{41.2} & 59.5          & 44.0          & \textbf{47.6} & 50.1          & \textbf{52.7} \\ \hline
\end{tabular}
\caption{\textbf{Experimental results of the scenario normal weather to foggy weather: Cityscapes→Foggy Cityscapes.} R34, R50 and CSP-D53 represent ResNet-34, ResNet-50 and CSP-Darknet53 respectively.}
\label{table_1}
\end{table*}

\subsection{Instance Feature Alignment and Consistency Loss}
Despite achieving domain adaptation between the backbone and encoder features through the alignment of local object-level and scene semantic features, there remains a bias towards the source domain in the decoder. Therefore, feature alignment in the decoder is essential. Considering that in DETR-style detectors, object queries are used as vector representations of query objects to describe objects, with each object query corresponding to a predicted box and containing its category and location information. Therefore, directly utilizing object query features for optimization through adversarial learning may interfere with detection performance, and is not an optimal solution. Drawing inspiration from previous work, 
we introduce an instance feature alignment module within the decoder, as illustrated in Figure~\ref{framework}c. 
We start by initializing a domain query and then extract instance-related features from the object queries using cross-attention, implemented with multi-head attention. Notably, this cross-attention layer does not share parameters with the cross-attention in the decoder, and the number of cross-attention layers is consistent with the number of decoder layers. We believe that this decoupled query can alleviate domain gap while maintaining the model's discriminative ability. The specific process can be described as follows:
\begin{equation}
    q_{i+1}=Linear(MultiHeadAttn(q_{i},k_{i},v_{i}))
\end{equation}
\begin{equation}
    \mathcal{L}^{D}_{adv}= -dlog(D_{d}(q_{i}) + (1-d)log(1-D_{d}(q_{i}))
\end{equation}
Herein, \(q\) denotes the domain query, \(i\) represents the layer index of the decoder (\(i=0,1,...,N\)), while \(k\) and \(v\) are both provided by the object query. \(D_d\) represents the discriminator within the decoder.

Unlike source domain data, target domain data cannot be optimized for detection loss using ground truth. Existing solutions typically adopt the mean-teacher framework, where pseudo-labels are obtained through a teacher model, and consistency constraints are utilized to enhance the learning of target domain images. To supervise target domain data, we further align the prediction distributions across different layers of the decoder, including category information, based on the existing adversarial loss. This approach enhances the model’s robustness without compromising its original detection capabilities. We employ the JS-divergence to measure the discrepancy in category distributions across different layers. The entire process is expressed as follows:
\begin{equation}
    \mathcal{L}_{cons}= \frac{1}{N}\sum_{i=1}^{N} JSD(\hat{y}_{cls},y^{i}_{cls})
\end{equation}
where $y^{i}$ represents the classification prediction result of the $i$-th layer. The reference output, denoted as $\hat{y}$, is obtained by averaging the predictions from all decoder layers, $JSD(\cdot)$ stands for the JS-divergence.

\subsection{Total Loss}
In summary, the overall optimization objective of the model is as follows:
\begin{equation}
\mathcal{L}_{total}=\mathcal{L}_{det}+\lambda_{1}\mathcal{L}^{B}_{adv}+\lambda_{2}\mathcal{L}^{E}_{adv}+\lambda_{3}\mathcal{L}^{D}_{adv}+\lambda_{4}\mathcal{L}_{cons}
\end{equation}
where $\mathcal{L}_{det}$ represent the detection loss, \(\lambda_{1}\), \(\lambda_{2}\), \(\lambda_{3}\), and \(\lambda_{4}\) represent the weight coefficients for different loss terms, respectively.

\begin{table*}[ht!]
\centering
\setlength{\tabcolsep}{8.0 pt} 
\renewcommand{\arraystretch}{1.2}
\begin{tabular}{l|c|cccccccc}
\hline
Method  & Backbone   & person        & rider         & car  & truck & bus  & mcycle & bicycle & mAP  \\ \hline
Source Only (Faster R-CNN)      & R50        & 28.8          & 25.4          & 44.1 & 17.9  & 16.1 & 13.9   & 22.4    & 24.1 \\
CR-SW~\cite{xu2020exploring}                    & V16          & 32.8          & 29.3          & 45.8 & 22.7  & 20.6 & 14.9   & 25.5    & 27.4 \\ \hline
Source Only (FCOS)       & R50        & 38.6          & 24.8          & 54.5 & 17.2  & 16.3 & 15.0   & 18.3    & 26.4 \\
EPM~\cite{hsu2020epm}         & R50        & 39.6          & 26.8          & 55.8 & 18.8  & 19.1 & 14.5   & 20.1    & 27.8 \\ \hline
Source Only (Def DETR)    & R50        & 38.9          & 26.7          & 55.2 & 15.7  & 19.7 & 10.8   & 16.2    & 26.2 \\ 
SFA~\cite{wang2021sfa}                        & R50         & 40.2          & 27.6          & 57.5 & 19.1  & 23.4 & 15.4   & 19.2    & 28.9 \\
AQT~\cite{huang2022aqt}                      & R50        & 38.2          & 33.0          & 58.4 & 17.3  & 18.4 & 16.9   & 23.5    & 29.4 \\
MTTrans\cite{yu2022mttrans}                 & R50         & 44.1          & 30.1          & 61.5 & 25.1  & 26.9 & 17.7   & 23.0    & 32.6 \\
MRT~\cite{zhao2023mrt}                 & R50        & 48.4          & 30.9          & 63.7 & 24.7  & 25.5 & 20.2   & 22.6    & 33.7 \\
BiADT~\cite{he2023bidirectional}                 & R50        & 42.0          & 34.5          & 59.9 & 17.2  & 19.2 & 17.8   & 24.4    & 32.7 \\

MTM~\cite{weng2024mean}                           & R50      & 53.7          & 35.1          & 68.8          & 23.0          & 28.8          & 23.8          & 28.0          & 37.3 \\ \hline
Source Only (RT-DETR)    & R34       & 47.5          & 37.4          & 69.7 & 27.0  & 28.0 & 24.9   & 32.3    & 38.1 \\
Source Only (RT-DETR)    & R50       & 51.6          & 40.3          & 71.5 & 31.6  & 35.3 & 29.0   & 34.3    & 41.9 \\
\rowcolor[HTML]{ECF4FF} RT-DATR(Ours)              & R34        & 54.0          & 44.3          & 73.0 & 33.6  & 35.6 & 30.4   & 34.9    & 43.7 \\
\rowcolor[HTML]{ECF4FF} RT-DATR(Ours)            & R50        & \textbf{57.0} & \textbf{45.9} & \textbf{74.5} & \textbf{36.4}  & \textbf{39.7} & \textbf{33.2}   & \textbf{37.9}    & \textbf{46.5} \\ \hline
\end{tabular}
\caption{\textbf{Experimental results of different methods for scene adaptation, Cityscapes → BDD100k daytime subset.} R34 and R50 represent ResNet-34 and ResNet-50, respectively. V-16 represents VGG-16. Def DETR represents Deformable DETR.}
\label{table_2}
\end{table*}

\section{Experiments}
We evaluated our approach on multiple cross-domain datasets, including weather adaptation (Cityscapes~\cite{cordts2016cityscapes} to Foggy Cityscapes~\cite{sakaridis2018foggycityscapes}), scene adaptation (Cityscapes to BDD100K~\cite{yu2020bdd100k}), artistic-to-real adaptation (Sim10K~\cite{johnson2016sim10k} to Cityscapes) and cross-camera adaptation (KITTI~\cite{geiger2012kitti} to Cityscapes). Ablation experiments were also conducted on the modules proposed in this paper. All results indicate that the approach presented in this paper is highly adaptable to various domain adaptation tasks across different scene datasets, significantly improving the algorithm's generalization performance.

\subsection{Datasets}
\noindent \textbf{Cityscapes to FoggyCityscapes.} Cityscapes is a semantic understanding dataset. It contains 2,975 training images and 500 validation images. FoggyCityscapes is a dataset generated by adding varying degrees of fog to the original Cityscapes images. The fog conditions are categorized into three levels: 0.005, 0.01, and 0.02, representing light, medium, and heavy fog, respectively. We selected the most challenging level, 0.02, to evaluate our experimental results on 8 common categories.

\noindent \textbf{Cityscapes to BDD100K.} BDD100K is a large-scale autonomous driving dataset that includes 70,000 training images and 10,000 validation images. We selected a subset that contains only daytime images to construct the adaptation benchmark. Therefore, we used 36,728 images for training and 5,258 images for validation. We chose Cityscapes as the source domain data and BDD100K as the target domain data, and evaluated 7 common categories.

\noindent \textbf{Sim10K to Cityscapes.} Sim10K is a dataset generated by a game engine, which contains 10,000 images and corresponding detection boxes. In the experiment, we selected Sim10K as the source domain data and Cityscapes as the target domain data. Only the ``car" category was included in the evaluation.

\noindent \textbf{KITTI to Cityscapes.} KITTI, another street scene dataset, differs from Cityscapes in that it originates from different cities and cameras. For our experiments, we utilized 7481 images as the training set, focusing solely on the car category that is common to both KITTI and Cityscapes.

\subsection{Implemention Details}
We selected RT-DETR as our base detector, utilizing ResNet-34 and ResNet-50 as backbone networks. The learning rate is set at 2e-4, and the batch size is 2. Training is performed on 8 A100 GPUs. For other settings, we follow the RT-DETR configuration, using the AdamW optimizer, 72 training epochs, and specifying an inference size of 640x640. Regarding loss weight coefficients, all are set to 1, except for $\lambda_{1}$, which is assigned a value of 1.5. For evaluation metrics, we report the Average Precision (AP) for each object category and the mean Average Precision (mAP) across all categories, with an IoU threshold of 0.5. Additionally, to ensure reliable initialization weights for the target domain, we pretrain the model on the source domain dataset.

\subsection{Comparison with Other Methods}
In this section, we compare the detection performance of RT-DATR with other current state-of-the-art methods (based on Faster R-CNN, FCOS, and DETR-style detectors) across four domain adaptation scenarios.

\begin{table}[ht!]
\centering
\setlength{\tabcolsep}{3.3 pt}
\renewcommand{\arraystretch}{1.2}
\begin{tabular}{c|ccc|c}
\hline
Method    & Faster RCNN & Def-DETR  & RT-DETR   & \multirow{2}{*}{mAP}  \\
Backbone  & VGG-16         & ResNet-50 & ResNet-34 &      \\ \hline
PLA       & \checkmark         &           &           & 37.3 \\
PLA       &             &           & \checkmark       & 41.8 \\
\rowcolor[HTML]{ECF4FF}LOFA(our) &             &           & \checkmark       & 45.6 \\ \hline
ILA       &             & \checkmark       &           & 36.8 \\
ILA       &             &           & \checkmark       & 38.1 \\
\rowcolor[HTML]{ECF4FF}IFA(our)  &             &           & \checkmark       & 40.9 \\ \hline
\end{tabular}
\caption{\textbf{Transfer the existing domain alignment method to RT-DETR and compare it with ours on Cityscapes→Foggy Cityscapes.} PLA represnet pixel-level alignment, ILA represent instance-level alignment.}
\label{table_5}
\end{table}

\begin{table}[ht!]
\centering
\setlength{\tabcolsep}{7.pt} 
\renewcommand{\arraystretch}{1.2}
\begin{tabular}{l|c|c}
\hline
Method        & Backbone  & car AP \\ \hline
Source Only (Faster R-CNN)        & R50         & 34.6   \\
DA-Faster~\cite{chen2018da-faster}       & R50        & 41.9   \\
GPA~\cite{xu2020gpa}               & R50       & 47.6   \\
KTNet~\cite{tian2021ktnet}              & R50         & 50.7   \\
PT~\cite{tian2021ktnet}               & R50         & 55.1   \\ \hline
Source Only (FCOS)          & R50     & 42.5   \\
EPM~\cite{hsu2020epm}             & R50         & 47.3   \\
SSAL~\cite{munir2021ssal}              & R50        & 51.8   \\ \hline
Source Only (Def DETR)       & R50         & 47.4   \\
SFA~\cite{wang2021sfa}          & R50         & 52.6   \\
AQT~\cite{huang2022aqt}            & R50         & 53.4   \\
DA-DETR~\cite{zhang2023da-detr}       & R50         & 54.7   \\
MRT~\cite{zhao2023mrt}           & R50       & 62.0   \\
MTM~\cite{weng2024mean}           & R50       & 58.1   \\ \hline
Source  Only (RT-DETR)      & R34        & 60.8   \\
\rowcolor[HTML]{ECF4FF} RT-DATR(Ours)  & R34        & \textbf{67.2}   \\ \hline
\end{tabular}
\caption{\textbf{Experimental results of the scenario synthetic scene
to real scene: SIM10k → Cityscapes.}}
\label{table_3}
\end{table}

\noindent \textbf{Motivation Explanation.} It is important to note that due to the architectural differences between one-stage and two-stage detectors, as well as the varying module designs in DETR-style detectors, directly transferring other methods to our baseline for comparison may not be entirely fair. Both AQT and ConfMix conducted ablation studies and discussions on the impact of different detector components on model performance. To further validate this limitation, we transferred the pixel-level alignment (PLA) method from ~\cite{zhou2022multi} and the instance-level alignment (ILA) method from AQT to RT-DETR and compared it with our proposed module. As shown in Table~\ref{table_5}, our proposed local object-level feature alignment (LOFA) module and instance feature alignment (IFA) module significantly outperform these coarse-grained baseline methods, which confirms the necessity and effectiveness of our work.

\noindent \textbf{Weather Adaptation.} We first  studied the adaptation from normal weather to foggy conditions based on the task Cityscapes→ Foggy Cityscapes. As shown in Table~\ref{table_1}, RT-DATR outperforms both the baseline and other state-of-the-art methods. For instance, with a ResNet34 backbone, RT-DATR achieves an mAP of 50.4\%, which is comparable to the metrics of models using a student-teacher framework, typically considered to yield better results. When using a ResNet50 backbone, RT-DATR attains an mAP of 52.7\%, surpassing the existing SOTA method, CAT, by 0.2\%.

\noindent \textbf{Scene adaptation.} Table~\ref{table_2} presents the results of the cross-scene adaptation experiment for the Cityscapes to BDD100K task. Following~\cite{xu2020exploring}, we report results for 7 common categories. It is evident that RT-DATR outperforms the most recent domain adaptive object detection methods, enhancing the mAP from 41.9\% achieved by the baseline model to 46.5\%. 

\begin{figure*}[h!]
\begin{center}
\includegraphics[scale=0.48]{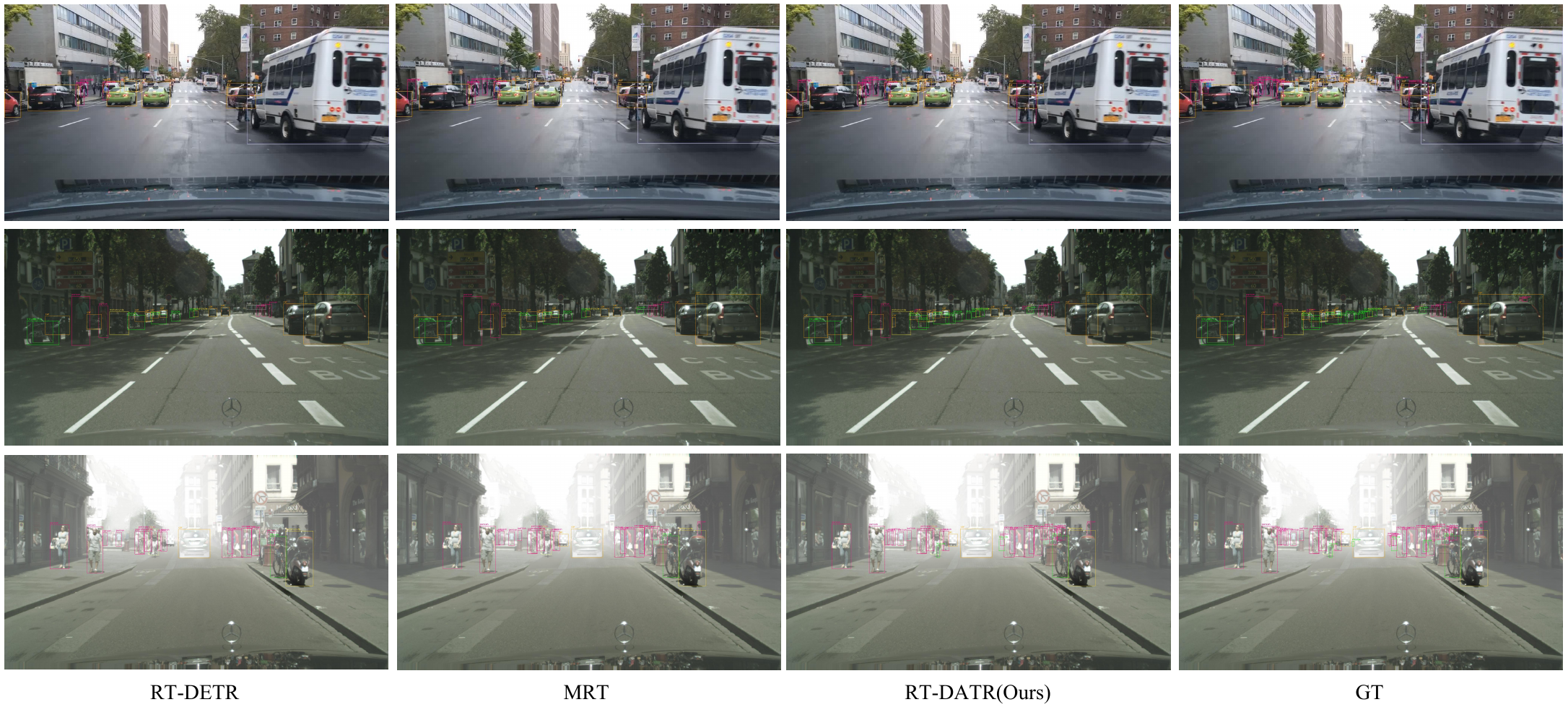}
\caption{Comparison of visualization results using different methods across various cross-domain detection datasets: the BDD100K dataset at the top, the Cityscapes dataset in the middle, and the Cityscapes Foggy dataset at the bottom.}
\label{res_show}
\end{center}
\end{figure*}

\begin{figure}[h!]
\begin{center}
\centerline{\includegraphics[scale=0.4]
{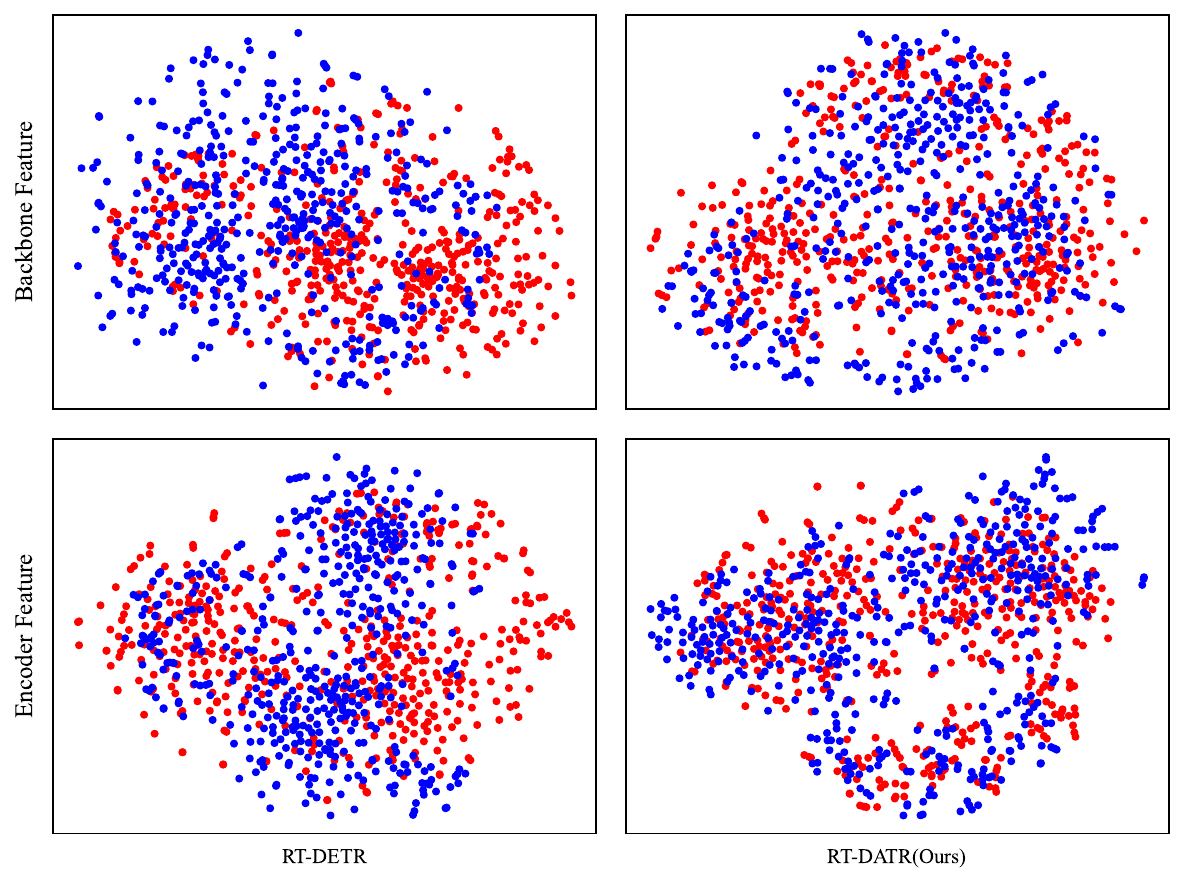}}
\caption{Feature visualization of the two domains on Cityscapes to Foggy Cityscapes by t-SNE. The blue and red points denotes source and target features respectively.}
\label{sne}
\end{center}
\vskip -0.2in
\end{figure}

 \noindent \textbf{Artistic to Real Adaptation.} Training a target detector from synthetic images and generalizing it to the real world can significantly enhance the diversity and utility of data sources. Table~\ref{table_3} presents the artistic-to-real adaptation experiment for the task Sim10K→Cityscapes. RT-DATR achieves an AP of 67.2\%, which is 5.2\% higher than the current state-of-the-art method, MRT. Furthermore, compared to CNN-based detectors, RT-DATR exhibits significant advantages in performance while also possessing real-time capabilities.

\begin{table}[]
\centering
\setlength{\tabcolsep}{7.pt} 
\renewcommand{\arraystretch}{1.2}
\begin{tabular}{l|c|c}  \hline
Method      & Backbone  & car AP \\ \hline
Source Only (Faster R-CNN)      & V16         & 38.5   \\
TIA~\cite{tian2019fcos}             & V16          & 44.0   \\ \hline
Source Only (Def DETR)      & R50         & 39.5   \\
SFA~\cite{wang2021sfa}            & R50        & 46.7   \\
DA-DETR~\cite{zhang2023da-detr}        & R50       & 48.9   \\\hline
Source Only (RT-DETR)     & R34        & 45.5   \\
\rowcolor[HTML]{ECF4FF} RT-DATR(Ours)  & R34       & \textbf{50.3}  \\ \hline
\end{tabular}
\caption{\textbf{Experimental results of the cross-camera adaptation: KITTI → Cityscapes.}}
\label{table_4}
\end{table}

\noindent \textbf{Cross-camera Adaptation.} Table~\ref{table_4} presents the cross-camera adaptation experiments for the task of KITTI→Cityscapes. It can be observed that RT-DATR achieves an AP of 50.3\% when utilizing a ResNet34 backbone, outperforming DA-DETR by +1.4\% and TIA by +6.3\%. These experimental results further demonstrate that the proposed method can generalize well to different domain adaptation tasks.

\noindent \textbf{Real-time Analysis.} It is worth noting that our endeavor does not lie in creating a novel real-time object detector. Rather, we strive to enhance the cross-domain detection capabilities of an existing transformer-based real-time detector, particularly emphasizing practical applications—a realm that remains largely uncharted in current research. The domain adaptation modules proposed in RT-DATR are training-only, ensuring that its inference latency remains consistent with that of RT-DETR. We conducted latency testing using TensorRT FP16 on a T4 GPU, with RT-DATR-R34 achieving a result of 6.3 ms (158 FPS). Not only is it faster, but its performance also surpasses that of the known CNN-based real-time domain adaptive detector, ConfMix. Moreover, we have achieved better results than other non-real-time cross-domain detectors under real-time conditions, and we believe that this real-time domain adaptive detection transformer has significant potential for widespread application.

\begin{table}[ht!]
\centering
\setlength{\tabcolsep}{4.0pt} 
\renewcommand{\arraystretch}{1.2}
\begin{tabular}{c|cccc|c}
\hline
Method & LOFA & SSFA & IFA & Cons loss & mAP  \\ \hline
 \multirow{8}{*}{RT-DETR-R34}   & x     & x    & x    &  x         & 35.6 \\
                              & \checkmark    &  x    &  x   &  x         & 45.6 \\
                              &  x    & \checkmark   &  x   &  x         & 42.8 \\
                              &   x   &  x    & \checkmark   &  x         & 40.9 \\
                              &  x    &   x   &  x   & \checkmark        & 37.5 \\
                              & \checkmark    & \checkmark   &  x   &      x     & 48.4 \\
                              & \checkmark    & \checkmark    & \checkmark   & x          & 49.5 \\
 & \checkmark    & \checkmark    & \checkmark   & \checkmark         & 50.4 \\ \hline
\end{tabular}
\caption{\textbf{Ablation study of RT-DATR on Cityscapes → Foggy Cityscapes.}}
\label{table_6}
\end{table}

\subsection{Ablation Study}
In Table~\ref{table_6}, we analyze the impact of four different components on the base detector (RT-DETR with a ResNet-34 backbone).  Firstly, it is clear that the LOFA module provides the most significant improvement, highlighting the importance of aligning object region features within the backbone network to represent domain-invariant features. Secondly, the SSFA module achieves a gain of +7.2\% compared to the baseline, indicating that, beyond local object-level feature alignment, scene semantic feature alignment is also indispensable in cross-domain detection. Furthermore, the IFA module contributes a 5.3\% improvement by aligning instance features, while the contribution of the consistency loss is comparatively smaller. Surprisingly, these methods are orthogonal, and their combined application leads to even greater benefits. This thoroughly demonstrates the effectiveness of our approach.

\noindent \textbf{Hyperparameter Study.} We conducted comparative experiments by setting different weight coefficients and discovered that the best performance was achieved when balancing the detection loss and adversarial loss, which is consistent with our prior knowledge.

\subsection{Visualization}
We use t-SNE~\cite{van2008visualizing} to visualize features from the backbone and encoder for the Cityscapes to Foggy Cityscapes adaptation. As illustrated in Figure~\ref{sne}, RT-DATR's features are less distinguishable between source and target domains compared to RT-DETR's more dispersed distribution. 
Furthermore, the encoder features in RT-DATR exhibit a higher degree of clustering compared to the backbone features, demonstrating that our method is capable of learning domain-invariant feature representations. 
Additionally, we compared the detection performance of the model, as shown in Figure~\ref{res_show}. RT-DETR and MRT exhibit similar performance, while RT-DATR outperforms MRT and RT-DETR,
achieving a higher precision and recall rate.

\section{Conclusion}
In this work, we propose a real-time domain-adaptive detection transformer, RT-DATR. We achieve local object-level and scene semantic feature alignment through adversarial feature learning, further mitigating the domain gap with domain queries, and enhance the model's robustness through consistency loss. Experimental results on multiple cross-domain benchmarks demonstrate the effectiveness of our method. Furthermore, considering the real-time capabilities and state-of-the-art performance of RT-DATR, we believe this can offer new insights and support to both the academic and industrial communities. 

{
    \small
    \bibliographystyle{main}
    \bibliography{main.bbl}
}

\end{document}